# Non-parametric spatially constrained local prior for scene parsing on real-world data

Ligang Zhang[*]

*School of Engineering and Technology, Central Queensland University, Brisbane, Australia*

l.zhang@cqu.edu.au

**Abstract:** Scene parsing aims to recognize the object category of every pixel in scene images, and it plays a central role in image content understanding and computer vision applications. However, accurate scene parsing from unconstrained real-world data is still a challenging task. In this paper, we present the non-parametric Spatially Constrained Local Prior (SCLP) for scene parsing on realistic data. For a given query image, the non-parametric SCLP is learnt by first retrieving a subset of most similar training images to the query image and then collecting prior information about object co-occurrence statistics between spatial image blocks and between adjacent superpixels from the retrieved subset. The SCLP is powerful in capturing both long- and short-range context about inter-object correlations in the query image and can be effectively integrated with traditional visual features to refine the classification results. Our experiments on the SIFT Flow and PASCAL-Context benchmark datasets show that the non-parametric SCLP used in conjunction with superpixel-level visual features achieves one of the top performance compared with state-of-the-art approaches.

## 1. Introduction

Scene parsing aims to assign every pixel of a query image to a correct semantic category, such as bus, road, sky and tree. It plays a central role in image content understanding and has the potential to support applications in a wide range of fields, such as object recognition, vehicle navigation, and risk identification. Although substantial progress has been made in recent years, scene parsing from complicated real-world data is still a very challenging task, due to substantial variability in both objects' properties (e.g., number, type, appearance, size and location), and environmental conditions (e.g., over-exposure, under-exposure, shining and illumination variations). Thus, developing techniques that are capable of robustly representing the most discriminative characteristics of all objects in the scene, while retaining high robustness against environmental variations has been a primary focus in this field.

---

[*] Corresponding Author.

The context about scene type and spatial layout presents very useful information in scene parsing. It is general knowledge that some objects (e.g., car and road, grass and cow, and sea and sand) are more likely to co-occur in the same scenes, whereas some other objects (e.g., sea and computer, and boat and train) are unlikely to co-exist in the same scene. In addition, the spatial layout also conveys important contextual information about the spatial location of objects within the scene. For instance, sky has a high probability of presenting in the top part of a city street scene, while road is likely to present at the bottom pat of the scene. Fig. 1 shows three examples of real-world scenes with such contextual information. Thus, prior context about object co-occurrence statistics and their spatial locations can be potentially used to improve scene parsing results.

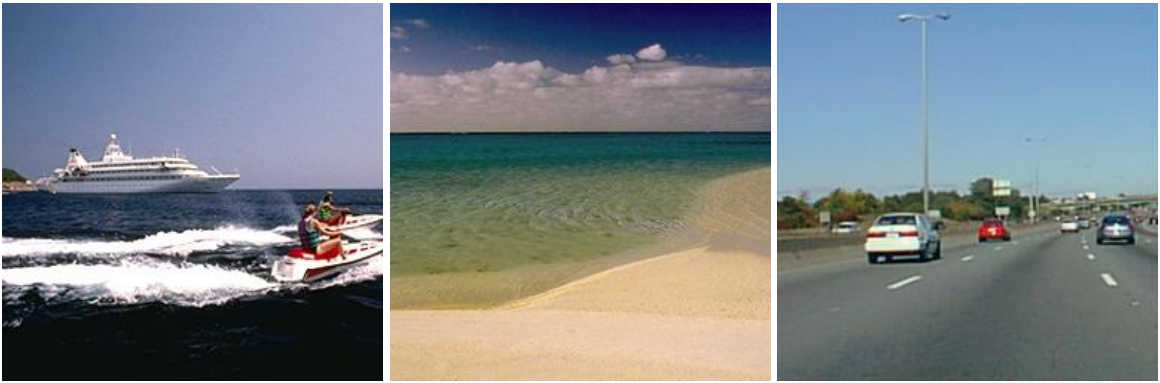

**Fig. 1.** Examples of real-world scenes. There exists prior context about co-occurrence and spatial locations of different objects.

Towards this aim, various efforts have been invested in the literature, resulting in various types of contextual features such as absolute location (Shotton et al., 2009), relative location (Gould et al., 2008), directional spatial relationships (Jimei et al., 2014), (Singhal et al., 2003), and object co-occurrence statistics (Micusik and Kosecka, 2009), as well as different types of graphical models such as Conditional Random Fields (CRFs) (Batra et al., 2008), (Lei and Qiang, 2010) and Markov Random Fields (MRFs) (Xiaofeng et al., 2012), and deep learning models such as Convolutional Neural Network (CNN), Fully Convolutional Network (FCN) (Shelhamer et al., 2017), DeepLab (Chen et al., 2018), Residual Network (ResNet) (Wu et al., 2019) and Context-contrasted Gated Boundary Network (CGBNet) (Ding et al., 2020). Although promising results have been achieved, most of those approaches still suffer from two drawbacks:

1) The training data are treated equally important in collecting contextual features or learning prediction models. However, in real-world scenarios, it is often the case that only a small portion of training data share similar characteristics to every individual query image. Rare classes also frequently present in realistic data and they may have big impact on scene parsing results (Shuai et al., 2018).

2) Limited capacity in capturing long- and short-range context, both of which present essential information for scene parsing. It is essential to create context-aware features to overcome big variations in scene content (Wang et al., 2017).

To address the drawbacks above, this paper presents the non-parametric Spatially Constrained Local Prior (SCLP) for scene parsing from complicated real-world data. It first retrieves a subset of similar training images to a query image and then collecting query-specific SCLP contextual features from the retrieved subset to represent more specific and useful prior context about inter-class correlations for the query image. The SCLP contextual features are further integrated with visual features in a decision-level fusion strategy to predict the class label of every superpixel in the query image. The power of non-parametric SCLP lies in its ability to represent both global and local context from the most similar training subset, separately for each query image, and further integrate contextual and visual features to improve the prediction results. It shows state-of-the-art performance on the SIFT Flow and PASCAL-Context benchmark datasets.

The main objectives of this paper are to 1) present a novel way of scene parsing on real-world data, which is particularly important for a wide range of practical applications. For instance, the proposed approach has already been applied to the segmentation and classification of objects (such as trees and grasses) for roadside fire risk prediction on the state roads in central Queensland, Australia; 2) validate the effectiveness of non-parametric contextual features, which are generated specifically for every query image based on a training subset that has similar scene context to the query image; and 3) provide a simple yet effective framework for the extraction and integration of both global and local context with visual features.

Compared with widely adopted deep learning techniques such as CNN and its variants, the proposed approach has three unique characteristics: 1) CNN utilizes convolutional and pooling layers to progressively extract more abstract patterns from the image and thus it emphasizes more on local context, while our approach extracts global and local context separately in two specifically-designed ways and thus it is able to consider both long- and short-range context. 2) CNN integrates visual and contextual features in a unified framework which often requires training a large amount of system parameters, while our approach provides a simple framework which involves only a couple of system parameters and is much easier to train and adjust. 3) CNN treats all images equally important during system training and thus it often has a difficulty of labelling rare classes due to insufficient training samples for those classes, while our approach collects query-specific contextual features from a retrieved training subset to represent more specific and effective context tailored to every query image including those of rare classes.

The rest of the paper is organized as follows. Section 2 discusses related work. Section 3 introduces the proposed non-parametric SCLP approach. The experiments are presented in Section 4 and finally Section 5 draws the conclusions.

## 2. Related Work

Early approaches (Shotton et al., 2009), (Bosch et al., 2007), (Kang et al., 2011) to scene parsing are often designed based on a set of visual features that are extracted at a pixel or region (e.g. patch and superpixel) level. The extracted features are then taken as inputs into a prediction algorithm to obtain the class label of every pixel or region. In these approaches, two critical tasks are: a) extracting visual features that are capable of capturing the most discriminative characteristics of all objects as much as possible, while retaining the background noise as little as possible, and b) the design of prediction algorithms to accurately discriminate one object from others based on the visual features. The most commonly used visual features include colour (e.g. RGB, Lab, HSV), texture, geometry, appearance, shape, etc., while the most popular algorithms include Support Vector Machine (SVM), Artificial Neural Network (ANN), Random Forest, and Adaboost. However, one major drawback of early approaches is that they have not considered the context information about the scenes with respect to the location and layout distributions of objects within the scenes. For instance, it is of a high probability that a cow and grasses co-exist close to each other in a 'cow' scene, while a cow is unlikely to present in a 'sea' scene.

To overcome the drawbacks in early approaches, many approaches have been proposed in recent decades to improve scene parsing results by more effectively utilizing the scene context. The context is often considered at two stages: feature extraction and label prediction. The typical types of contextual features include absolute location (Shotton et al., 2009), relative location (Gould et al., 2008), directional spatial relationships (Jimei et al., 2014), (Singhal et al., 2003), and object co-occurrence statistics (Micusik and Kosecka, 2009). Because these features often represent the prior information about the spatial distributions of objects or the inter-object spatial correlations, they have demonstrated a powerful role in further improving the results of visual features. As for label prediction, two most popular prediction algorithms for considering scene context are CRFs (Batra et al., 2008), (Lei and Qiang, 2010) and MRFs (Xiaofeng et al., 2012), which refine the prediction results by enforcing the spatial consistency of class labels between neighbouring pixels or regions. Although CRF and MRF have led to substantial improvements, they primarily focus on the consistency within local neighbourhoods and thus they consider the context only to a limited range. In realistic data, the context information can present all over the whole scene. To consider a larger range of context, various types of extensions have also been proposed, including hierarchical CRF (Lempitsky et al., 2011), stacked hierarchical model (Munoz et al., 2010), recursive context propagation network (Sharma et al., 2014), pylon model (Lempitsky et al., 2011), histograms in superpixel neighbourhoods (Fulkerson et al., 2009), etc. A representative work is (Cheng et al., 2019), which adopted random forest classifiers and convolutional networks to improve the confidence of initial semantic segmentation. A global optimization was then performed using a CRF model to refine semantic labels. The work focuses on refining the predicted probabilities of semantic segmentation algorithms, while our work in this paper is to exploit the effectiveness of global and local context.

Recent advances also demonstrated highly competitive results of scene parsing using deep learning techniques, such as Convolutional Neural Networks (CNNs) (Farabet et al., 2013), Multi-scale CNNs (Yu and Koltun, 2015), Fully Convolutional Networks (FCNs) (Shelhamer et al., 2017), Contextual Hierarchical Models (CHMs) (Seyedhosseini and Tasdizen, 2015), Deep Parsing Network (DPN) (Liu et al., 2015), DeepLab (Chen et al., 2018), ResNet (Wu et al., 2019) and CGBNet (Ding et al., 2020). A major advantage of deep learning techniques lies in the use of a single multi-layer architecture to effectively extract and seamlessly integrate both visual and contextual features, and further combine them with a prediction layer to derive the final class labels. A representative deep network is the DeepLab (Chen et al., 2018), which is essentially a cascade of two very well-established modules - Deep Convolutional Neural Network (DCNN) and CRF. The DeepLab has specifically designed mechanisms to capture larger context by resampling the feature layer at multiple scales prior to convolution and applying upsampled filters in the last few convolutional layers of the DCNN. Another most recent deep network is the CGBNet (Ding et al., 2020), which encodes informative context into local features and selectively fuses the segmentation results from different-scales of features. However, there are little efforts on designing context-adaptive features capable of automatically adapting to both the local and global contextual properties in every new test image.

In (Zhang et al., 2016), we first proposed the SCLP to represent the contextual prior information about the scenes. The SCLP collects inter-object co-occurrence statistics between spatial blocks and between neighbouring superpixels from the training dataset to refine the results of visual features. The SCLP used in conjunction with traditional visual features has demonstrated promising performance on scene parsing. However, one drawback of the SCLP is that it is collected from all training data without taking into account the content and contextual similarity of each training data to the query image. In realistic conditions, it is often the case that only a small portion of training data show very similar content and context to a query image. Thus, it would be beneficial to adopt a non-parametric approach to collect an individual SCLP specifically for each query image. It is anticipated that the resulting non-parametric SCLP can more accurately reflect the prior contextual information about the query image. This forms the main motivation of the proposed non-parametric SCLP.

The non-parametric SCLP is also inspired by and built upon existing non-parametric approaches (Tighe and Lazebnik, 2010), which have demonstrated promising performance on scene parsing. A similar work to ours is (Hung et al., 2017), which learned spatial and global prior information by retrieving the K-nearest annotated images from the training set, and then encoded the prior information into the process of image segmentation. However, the work (Hung et al., 2017) has primarily considered the global context. Another similar work is (Zhang et al., 2019), which modelled the scene context using co-occurrent features for a given target object. The probability distribution of the co-occurrent features conditioned on the target was predicted using a Mixture of Softmaxes and further aggregated to capture co-occurrent context information across the scene. However, the work mainly focused on the global context and did not consider context-adaptive features specifically for every test image. Different from those works, the proposed non-parametric SCLP is designed to capture both long- and short-range context of a query image from the retrieved data subset.

## 3. Proposed Non-parametric SCLP Approach

Fig. 2 shows an overview of the system framework for the proposed non-parametric SCLP approach for scene parsing. For a given query image, the approach processes it in three main processing streamlines: 1) non-parametric SCLP extraction and prediction; 2) visual feature extraction and prediction; and 3) prediction fusion and voting.

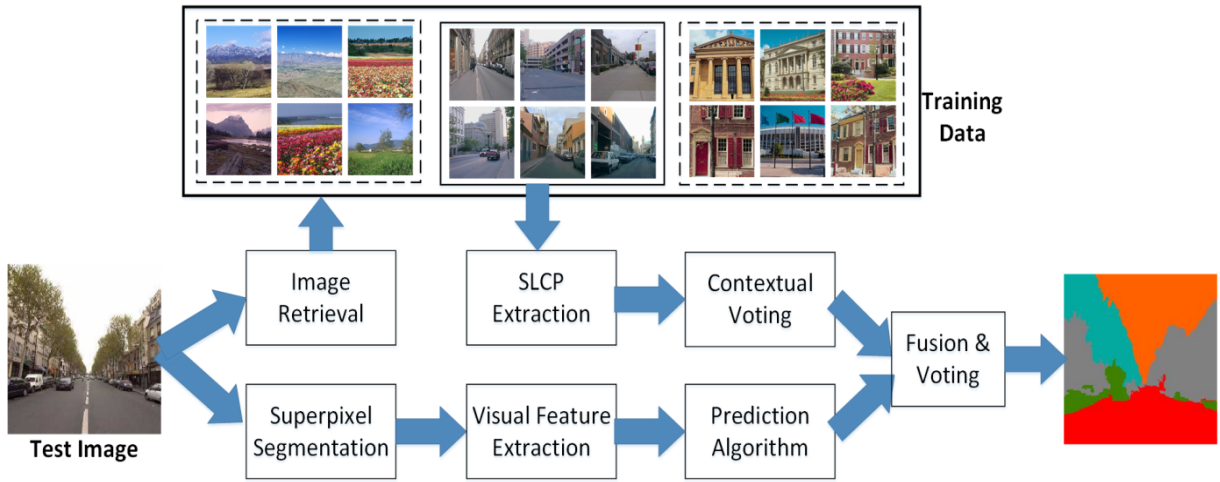

**Fig. 2.** Framework of the proposed non-parametric SCLP approach for scene parsing.

For non-parametric SCLP extraction, an image retrieval process is introduced to retrieve a subset of images from the training data that have similar content and contextual information to the query image. The SLCP is then collected from the retrieved training subset and further taken as input into a contextual voting process to obtain the contextual probabilities of every pixel belonging to all object categories. As for visual feature extraction, both training and the query images are first segmented into a set of superpixels $S = \{S_1, S_2, ..., S_L\}$, where $L$ is the total number of superpixels and $S_l$ indicates the $l^{th}$ superpixel, with homogeneous visual appearance using a fast graph-based image segmentation algorithm (Felzenszwalb and Huttenlocher, 2004). A set of visual features is then extracted for every superpixel. The visual features are further fed into an ANN to predict the visual probabilities of every superpixel belonging to all object categories. The visual and contextual probabilities are then fused and go through a majority voting process to obtain the class label for every superpixel, such as road, car, tree, building and sky. Details of those processing steps are described below.

*3.1 Retrieval of Similar Training Images for Every Test Image*

For a given query image, the first step of the proposed approach is to retrieve a subset of images from the training data, which share similar visual characteristics to the query image. It is expected that the image subset contains similar types of scene content and context to the query image, with respect to object categories and their spatial layouts. As an example shown in Fig 2, given a query image of a 'city road' scene, the retrieved subset is anticipated to be 'city road' scenes rather than 'grass' or 'building' scenes.

To achieve this goal, we use four types of global image features as in (Tighe and Lazebnik, 2010): spatial pyramid represented by SIFT dictionaries, GIST, Tiny image, and RGB color histogram. For each type of features, all training images are ranked based on their Euclidean distances from the query image, and the top $\alpha$ images with the smallest distances are kept in the final retrieval set. For four types of features, we can obtain the resulting retrieval set $R = \{R_1, R_2, \dots, R_p\}$, where *p=4\*α* is the total number of retrieved images and $R_p$ indicates the $p^{th}$ image. One benefit of retrieving an image subset separately for each feature is that the retrieved images are able to reflect different characteristics of the query image and thus they retain a high diversity of the scene content. We anticipate that the retrieved image subset enables more accurate contextual feature extraction, particularly for the situations where the query image has big variations from the training data.

*3.2 Extraction of Non-parametric SCLP*

Given the retrieved image subset, we proceed to extract the SCLP features from them. The SCLP features essentially capture two types of contextual information: inter-object correlations in spatial image blocks and object co-occurrences in neighbouring superpixels. They reflect the long- and short-range prior contextual information respectively learned from retrieved training images. The calculation of them is described as follows.

1) *Inter-object correlations in spatial blocks (global SCLP)*. For the $p^{th}$ image $R_p$ in the retrieved subset, it is divided into a number of equal spatial blocks $\{R_p = \bigcup(B_k), k = 1, 2, \dots, K\}$ and $K$ is the number of blocks. For simplicity, this paper uses blocks that have the same height and width. Fig. 3 illustrates an example image and its corresponding spatial blocks. Because the proposed approach takes superpixels as basic processing unit, all pixels within the same superpixel should be assigned to only one spatial block. In this paper, we assign every superpixel to a unique block based on its centroid. Let assume the $l^{th}$ superpixel $S_l$ is assigned to the $k^{th}$ block $B_k$ and $\mathbb{C}(S_l)$ is the total number of pixels in $S_l$, all pixels $p_i$ within $S_l$ are assigned to the block $B_k$, i.e. $p_i \to B_k$, $i \in \{1, 2, \ldots, \mathbb{C}(S_l)\}$ and $S_l \in B_k$. Fig. 3 provides an example superpixel to show that all pixels within the superpixel are assigned to the $8^{th}$ block based on their centroid. Let us assume an image pixel with a class label $\acute{c}$ occurs in a block $B_{k_1}$, a matrix $\mathcal{M}_{c|\acute{c}}(k_1, k_2)$, $k_1 \neq k_2$ stands for the probability that a pixel with a class $c$ occurs in a block $B_{k_2}$ and $\mathcal{M} \in \mathcal{R}^{M*M*K*(K-1)}$, where $M$ indicates the number of all classes. The matrix can be obtained for each pair of blocks in all images in the retrieved subset, and is further normalized to ensure a conditional probability distribution over all classes, i.e. $\sum_{c=1}^{M} \mathcal{M}_{c|\acute{c}}(k_1, k_2) = 1$. The global matrices reflect the global support from other blocks on the class label of pixels within a certain block.

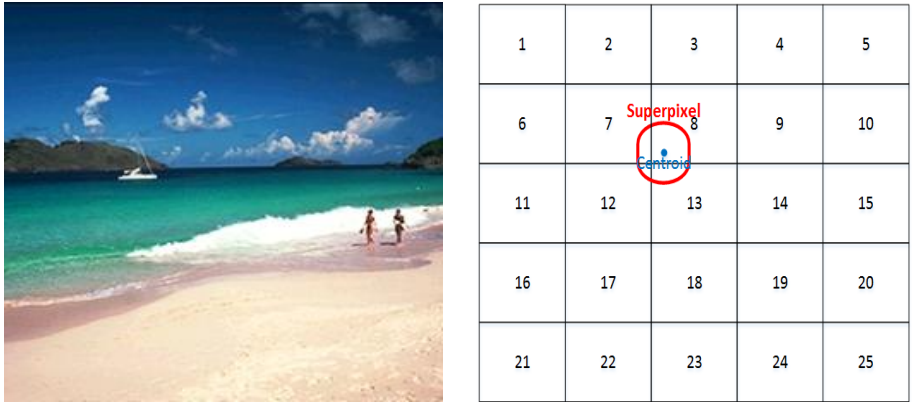

**Fig. 3.** Illustration of assigning a superpixel to an image block. The example image (left) is divided into 25 blocks (right). All pixels within the example superpixel (in red) is assigned to the $8^{th}$ block based on their centroid (in blue), although a proportion of superpixels is also located within the $7^{th}$, $12^{th}$, and $13^{th}$ blocks (best viewed in colour).

2) *Object co-occurrence in neighbouring superpixels (local SCLP)*. For the $p^{th}$ image $R_p$ in the retrieved subset, assume it is segmented into a set of superpixels $\{R_p = \cup(S_l), l = 1, 2, \ldots, L\}$, and $L$ is the number of superpixels. For the $l^{th}$ superpixel $S_l$ with a class label $\acute{c}$, a matrix $\overline{\mathcal{M}}_{c|\acute{c}}(S_l, S_p)$ is formed to represent the probability that one of its adjacent superpixel $S_p$ with a class $c$ and $\overline{\mathcal{M}}_{c|\acute{c}}(S_l, S_p) \in \mathcal{R}^{M*M}$. The matrix is calculated for each pair of neighboring superpixels in all images in the retrieved subset, and further normalized in the same way as the global matrix. The local matrices reflect the local support from neighbouring superpixels on the class label of pixels within a certain superpixel. Note that it is assumed all pixels within every superpixel have the same class label. Although this assumption is not strictly true for all real-world situations, it is generally acceptable in most of real-life scene data.

*3.3 SCLP-Based Class Probability Prediction (Contextual Voting)*

The global and local matrices contain prior contextual information regarding the likelihood of two objects co-occurring in the same scene. For a new query image $I$ and its segmented superpixels $\{I = \cup(S_l), l = 1,2,...,L\}$, we can obtain the likelihoods of each superpixel $S_l$ belonging to each of all object categories $C = \{c_i | i = 1,2,...,M\}$, based on the global and local matrices. This is completed by casting a vote from a superpixel to the rest of all superpixels, and then accumulating the votes to form the final class probabilities.

For a superpixel $S_l$ in a block $B_{k_2}$, let $\hat{S} = \{S_q | q = 1,2,...,Q\}$ be the set of superpixels in all the rest of blocks $\cup(B_{k_2}) \ k_1 \neq k_2$ and $S_q$ be the $q^{th}$ element of $\hat{S}$. The $S_l$ receives Q votes from all superpixels in $\hat{S}$, $S_q \in \hat{S}$, based on the global SCLP matrix:

$$V^{gobal}(C|S_l) = \sum_{k_1 \neq k_2; S_q \in \acute{c}} w_q * \mathcal{M}_{c|\acute{c}}(k_1, k_2) \qquad (1)$$

where $V^{gobal}$ indicates the votes received based on the global SCLP matrix, $w_q = P^{visual}(\hat{c}|S_l) * \mathbb{C}(S_q)$ is a weight given to the vote from the superpixel $S_q$, $\mathbb{C}(S_q)$ is the total number of pixels in $S_q$, and $P^{visual}(\hat{c}|S_l)$ is the probability of the most probable class $\hat{c}$ predicted using visual features of $S_l$ (details are in Section 3.4). The $V^{gobal}$ is then normalized to a probability $P^{gobal}$ by dividing the total number of votes for all $M$ classes:

$$P^{gobal}(C|S_l) = V^{gobal}(C|S_l) / \sum_{c=0}^{M} V^{gobal}(c|S_l) \qquad (2)$$

In a similar way, let $\bar{S} = \{S_p | p = 1,2,...,P\}$ be the set of adjacent superpixels of $S_l$ and $S_p$ be the $p^{th}$ member of $\bar{S}$, the $S_l$ receives $P$ votes from all its adjacent neighbours based on the local SCLP matrix:

$$V^{local}(C|S_l) = \sum_{1<p<P; S_p \in \acute{c}} w_p * \bar{\mathcal{M}}_{c|\acute{c}}(S_l, S_p) \qquad (3)$$

where $V^{local}$ indicates the votes received based on the local SCLP matrix, and $w_p$ is the weight defined in the same way as $w_q$. The $V^{local}$ is further converted into a probability $P^{local}$ in the same way as $P^{gobal}$.

*3.4 Visual Feature-Based Class Probability Prediction*

Visual feature-based prediction aims to obtain the probability of each superpixel belonging to each class category using solely superpixel-level visual features. The features are adopted from (Tighe and Lazebnik, 2010) and include 537 color, geometric and texture features such as histogram of RGB colors, textons and SIFT descriptors over the superpixel region. The minimum Redundancy Maximum Relevance (MRMR) algorithm (Hanchuan et al., 2005) is further used to reduce the dimension of features to 50, for each class separately.

The MRMR algorithm was designed to select a subset of features which has the least similarity between them (redundancy), but the highest correlation with ground truth classes (relevance). The reduced feature set is then fed into a feedforward ANN algorithm to obtain the visual probability of each superpixel: $P^{visual}(C|S_l)$, where $C = \{c_i | i = 1, 2, \ldots, M\}$ indicates the class labels and $M$ indicates the number of classes. Note that, except ANN, any algorithm with probability prediction capacity can be used here, such as logistic regression and support vector machine.

*3.5 Fusion of Visual and Contextual Probabilities*

For the $l^{th}$ superpixel $S_l$ in a query image, we can obtain its corresponding global contextual probability $P^{global}(C|S_l)$, local contextual probability $P^{local}(C|S_l)$, and visual probability $P^{visual}(C|S_l)$. To predict the class label of $S_l$, we first adopt a decision-fusion strategy to combine the three probabilities into a single probability:

$$P(C|S_l) = w^c + w^g * P^{gobal}(C|S_l) + w^l * P^{local}(C|S_l) + w^v * P^{visual}(C|S_l) \quad (4)$$

where, $w^c$ is a constant value, $w^g$, $w^l$ and $w^v$ are weights given to $P^{gobal}(C|S_l)$, $P^{local}(C|S_l)$ and $P^{visual}(C|S_l)$ respectively.

Finally, a majority voting strategy is employed to assign the superpixel $S_l$ to the $i^{th}$ class $c_i$ which owns the highest probability across all classes:

$$S_l \in \bar{c} \text{ if } P(\bar{c}|S_l) = \max_{1 < i < M} P(c_i|S_l) \quad (5)$$

**4. Experimental Results and Analysis**

This section evaluates the proposed non-parametric SCLP approach on the widely used SIFT Flow and a more recent PASCAL-Context dataset. We compare the classification performance obtained with different numbers of retrieved images and of rare classes on the SIFT Flow dataset. The performance is also compared with those of state-of-the-art approaches on both datasets.

*4.1 Evaluation Datasets and System Parameters*

1) *Evaluation datasets*. The first dataset used for the experiments is the widely used SIFT Flow dataset, which comprises of 2,688 images and most of them are collected from real-world scenarios. The primary scenes include street, beach, mountain, field, building, etc. and the top 33 class categories with the most 10 labelled pixels are chosen. The images are pixel-wisely labelled by LabelMe users and an additional 'unlabeled' category is also included to indicate pixels that are not labelled or labelled as other categories. Following the common train/test data split (Ce et al., 2009), we use 2,488 and 200 images for training and evaluation respectively.

The second dataset is the PASCAL-Context dataset (Mottaghi et al., 2014), which includes pixel-wise annotations of 540 class categories for the 10,103 *trainval* images of the PASCAL VOC 2010 challenge. Following previous work (Mottaghi et al., 2014), (Shelhamer et al., 2017), the most frequent 59 classes are used in the classification task here, and the training and validation are based on the *train* and *val* sets respectively.

2) *System parameters*. For the experiments, the initial number of spatial blocks is set to 6×6=36. The graph-based image segmentation algorithm is implemented based on the following initial parameters: $\sigma = 0.8$, $min = 100$, $k = 200 \times \max(1, sqr(D_I/640))$, and $D_I$ is the larger one between height and width of an image *I*. The algorithm used for predicting visual probability is a three-layer feedforward ANN comprising of 16 hidden neurons. For the MRMR feature selection algorithm, the continuous input features are converted into three discrete values [1, 0, -1] based their mean value $\mu$ and standard deviation $std$. The two thresholds are $\mu + w * std$ and $\mu - w * std$, respectively, where *w* is set to 0.5.

3) *Evaluation metrics*. Following previous works (Tighe and Lazebnik, 2013), (Shelhamer et al., 2017), we use four common metrics for performance evaluation:

i) global accuracy: $\sum_i^M n_{ii} / \sum_i^M t_i$, which is the ratio of correctly predicted pixels to the total pixels in all test images;

ii) class accuracy: $\frac{1}{M}\sum_i^M n_{ii}/t_i$, which is the average pixel accuracy over all categories;

iii) mean IU: $\frac{1}{M}\sum_i^M n_{ii}/(t_i + \sum_j^M n_{ji} - n_{ii})$, where is the region Intersection over Union (IU);

iv) frequency weighted IU (FW IU): $(\sum_k^M t_k)^{-1} \sum_i^M t_i n_{ii}/(t_i + \sum_j^M n_{ji} - n_{ii})$, where $n_{ij}$ indicates the number of pixels of the *i*[th] class predicted as the *j*[th] class, $t_i = \sum_j^M n_{ij}$ is the number of pixels of the *i*[th] class, and *M* is the number of classes.

## 4.2 Performance vs. System Parameters

This part investigates the impact of two system parameters, including number of retrieved images and number of rare classes, on the scene labelling performance of the proposed approach. The results on the SIFT Flow dataset are used for the analysis here.

1) *Performance vs. number of retrieved images*. A key parameter of the proposed approach is the number of images retrieved for every query image, which directly impacts the results of global and local SCLP, and thus impacts the accuracy on the query image. Fig. 4 shows the global and average class accuracies obtained using a different number of images retrieved from the training data on the SIFT Flow dataset. As can be seen, using a larger number of retrieved images leads to consistent improvements to both global and class accuracies when the number is smaller than 100, after which the performance tends to level off. When only top 100 retrieved images are used, the global and class accuracies are 86.4% and 38.1% respectively. It is interesting to note that using a small number of retrieved images is able to achieve quite promising results. For instance, when only top 10 retrieved images are used, the global and class accuracies are still higher than 80% and 32% respectively. The results imply that a small number of most similar images can largely determine the performance of global and local SCLP on the SIFT Flow dataset.

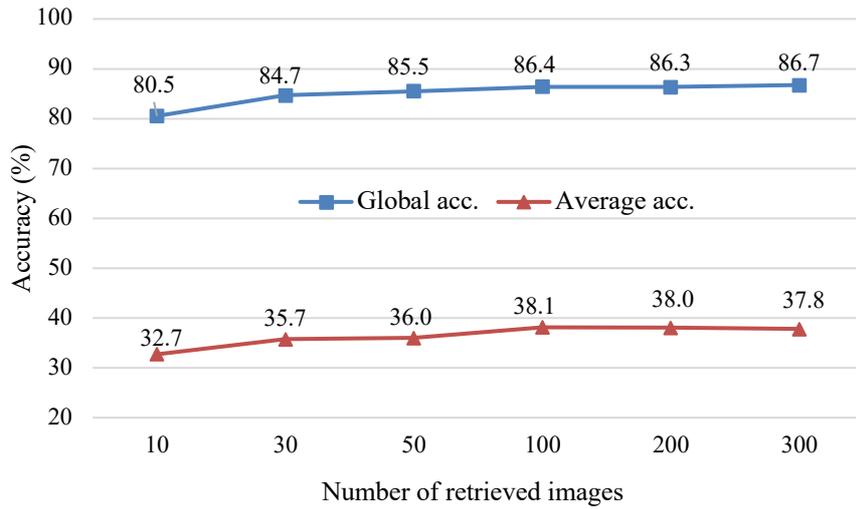

**Fig. 4.** Global and class accuracies vs. number of retrieved images on the SIFT Flow dataset.

2) *Performance vs. number of rare classes*. One appealing fact of real-world datasets such as SIFT Flow is that their pixel distribution is often heavily tailored to several frequently occurring classes, and there are some rare classes with only very limited number of pixels. Fig. 5 shows the pixel distributions of 33 classes on the SIFT Flow dataset. The proposed approach has the advantage of handling this situation by retrieving a set of similar images separately for each rare class and utilizing the SCLP information from the retrieved image set to refine the classification results. To demonstrate this advantage, we retrieve similar image subsets separately for the top 6, 9, and 14 rare classes as compared to using all 33 classes. The results are shown in Table 1. It can be seen that retrieving images for only rare classes achieves higher global and class accuracies than retrieving images for all classes. This is an interesting and important finding as it indicates it is not always necessary to perform time-consuming image retrieval for all classes on real-world datasets and focusing on rare classes is able to achieve higher accuracy with less processing time. The highest global accuracy is 87.2% and class accuracy is 42.5% when the top 14 and 9 rare classes are considered respectively. However, it is noted that the difference on global accuracy is relatively smaller than that on class accuracy using different numbers of top rare classes.

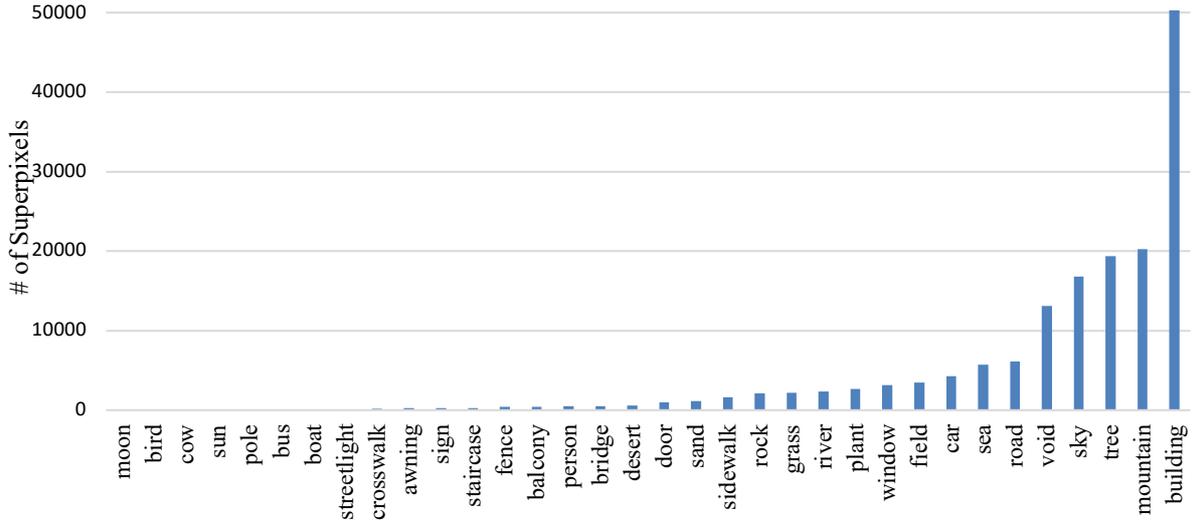

**Fig. 5.** Number of segmented superpixels for objects on the SIFT Flow dataset. Note the majority of the superpixels are from several dominant objects, and there are many rare objects with only a small number of superpixels.

**Table 1**
Global and class accuracies (%) vs. number of top rare classes.

|  | Number of top rare classes | | | |
| --- | --- | --- | --- | --- |
|  | 6 | 9 | 14 | 33 (all) |
| **Global Acc.** | 87.1 | 86.8 | 87.2 | 86.4 |
| **Class Acc.** | 40.8 | 42.5 | 41.9 | 38.1 |

3) *Performance vs. number of spatial blocks*. One key parameter for obtaining global SCLP features is the number of spatial blocks, which directly determines the size of image regions from which object co-occurrence frequency is calculated and global SCLP is learnt. Fig.6 shows the impact of the number of spatial blocks on the global and class accuracies. As can be seen, the proposed approach achieves peak performance of 87.2% and 43.0% for global and class accuracies respectively when 16 (i.e., 4×4) blocks are used. Using a larger number of spatial blocks does not lead to a better performance, which is within our expectation as the spatial block is designed to capture long-range label dependencies of objects and using too many blocks may make it difficult to learn sufficient label dependencies for all block pairs. Similarly, using a smaller number (e.g., 4) of spatial blocks may not capture representative features for distinguishing some objects because those objects may co-occur in the same block. However, the performance fluctuations are still relatively small and within an absolute range of 2%, implying limited impact of using different number of spatial blocks on the overall performance.

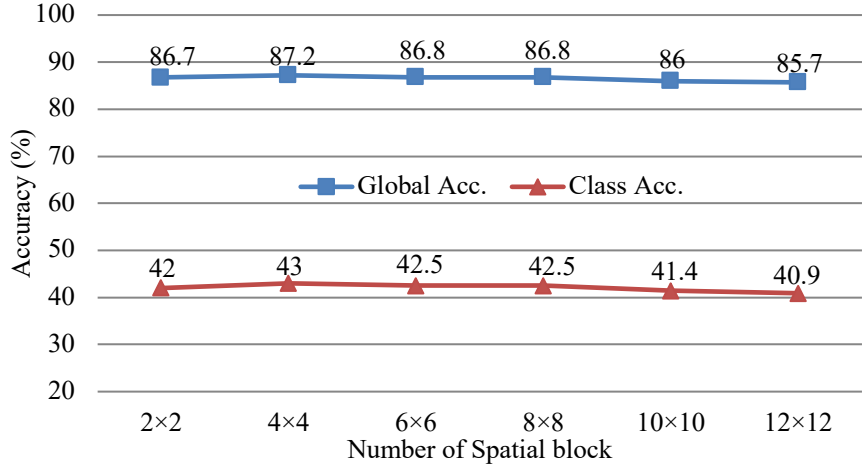

**Fig. 6.** Global and class accuracies vs. number of spatial blocks on the SIFT Flow dataset.

4) *Performance vs. number of superpixels*. Since superpixels are the basic unit for extracting visual, local SCLP and global SCLP features, the number of superpixels has direct impact on the performance. Unfortunately, there is no direct way of evaluating the impact because the image segmentation algorithm often produces substantially different number of superpixels for different images due to a great variety of scene content, type and layout. This is also the case even when the same set of algorithm parameters is used for all images. To provide a rough idea about the impact, we run the image segmentation algorithm with different sets of parameters and then get the average number of superpixels in all images. Table 2 shows the relationships between accuracies and the average number of superpixels. The highest global accuracy of 87.2% and highest class accuracy of 43.0% are obtained when an average of 158 superpixels is used. Similar to the results using different number of spatial blocks, using either a small or a large amount of superpixels leads to decreased global and class accuracies. This is particularly evident when more than 300 superpixels are used. One possible reason is that it becomes harder to learn discriminative features from small superpixels. The above results indicate that choosing a suitable number of superpixels has significant impact on the performance.

**Table 2**
Global and class accuracies (%) vs. average number of superpixels.

|  | **Average number of superpixels** | | | | |
| --- | --- | --- | --- | --- | --- |
|  | 64 | 102 | 158 | 253 | 356 |
| **Global Acc.** | 80.3 | 86.9 | 87.2 | 84.9 | 72.7 |
| **Class Acc.** | 37.8 | 42.6 | 43.0 | 39.0 | 34.6 |

5) *Computational time performance*. The whole system was implemented under a Matlab platform using a Macbook laptop with a configuration of 1.8 GHz Intel Core i5 processor and 4GB memory. The average computational time required for every test image is 0.78 second per image. We anticipate that the computational time could be significantly improved if a more efficient programming language was used such as C++. It should be noted that the time is dependent on a couple of factors, such as image resolution, dataset size, number of retrieved images, number of top rare classes, number of superpixels, and number of blocks.

*4.3 Performance Comparisons with State-of-the-Art Approaches*

1) *Performance comparisons on the SIFT Flow dataset.* Table 3 compares the proposed approach with state-of-the-art approaches on the SIFT Flow dataset. It can be seen that the proposed non-parametric SCLP achieves competitive performance of global and class accuracies compared to existing approaches. The proposed SCLP has a class accuracy of 43.0%, which is relatively lower than those in other top-performing approaches such as (Shuai et al., 2018) and (Ding et al., 2020). This is likely due to only a small number of training data available for some rare classes on the SIFT Flow dataset. However, this has significant impact on training subset retrieval and SCLP feature extraction for those rare classes in the proposed approach.

To provide more insights into possible improvements to the proposed approach, we do more analysis on the similarities and differences between the proposed approach and other top-performing approaches. The CGBNet in (Ding et al., 2020) has the highest global accuracy of 89.7%, which is 2.5% higher than the proposed approach. Similar to our approach, the CGBNet generates context-contrasted local features to make use of the informative context. A unique feature of the CGBNet is that it has a scale-selection step to selectively fuse the segmentation results from different-scale features. The results may indicate the importance of considering multi-scale features in our future work. Another top performer is the Directed Acyclic Graph - Recurrent Neural Network (DAG-RNN) in (Shuai et al., 2018), which explicitly aggregates contextual information into local features extracted using a pre-trained CNN. Similar to our approach, the DAG-RNN was designed to model the contextual dependencies between image regions and then inject the informative context to enhance the representative capability of local features. Different from the proposed approach, the DAG-RNN primarily focuses on local context and does not consider adaptive context for every query image. A main advantage of the DAG-RNN is that it employs a class-weighted function to distribute reasonably higher attention weights to infrequent classes based on their rareness magnitude, which significantly improves the overall performance.

Fig. 7 visually illustrates the scene parsing results on several sample images from the SIFT Flow dataset. These samples contain different types of scene content (e.g., city street vs. forest, and building vs. mountain) and have big variations in environmental conditions such as different lighting conditions. From these samples, we can see that the proposed non-parametric SCLP demonstrates a good overall performance in labelling each test image pixel into a correct class label. The proposed approach also shows strong robustness under different environmental factors such as shadows and illumination variations. We also can observe that there are some small regions or objects that are misclassified by the proposed approach primarily due to their small sizes, which increase the difficulty in learning effective SCLP contextual features.

**Table 3**

Performance (%) comparisons with state-of-the-art approaches on the SIFT Flow dataset (33 classes).

| Approach | Global Acc. | Class Acc. |
|---|---|---|
| (Farabet et al., 2013) | 78.5 | 29.6 |
| (Tighe and Lazebnik, 2013) | 78.6 | 39.2 |
| (Singh and Kosecka, 2013) | 79.2 | 33.8 |
| (Jimei et al., 2014) | 79.8 | 48.7 |
| (Sharma et al., 2014) | 79.6 | 33.6 |
| (Najafi et al., 2015) | 76.6 | 35.0 |
| (Nguyen et al., 2015) | 78.9 | 34.0 |
| (Sharma et al., 2015) | 80.9 | 39.1 |
| (Shelhamer et al., 2017) | 85.9 | 53.9 |
| (Lu et al., 2017) | 77.9 | 42.8 |
| (Shuai et al., 2018) | 87.8 | 57.8 |
| (Ye et al., 2018) | 82.1 | 58.8 |
| (Ates and Sunetci, 2019) | 88.7 | 55.2 |
| (Ding et al., 2020) | 89.7 | 58.5 |
| Non-parametric SCLP | 87.2 | 43.0 |

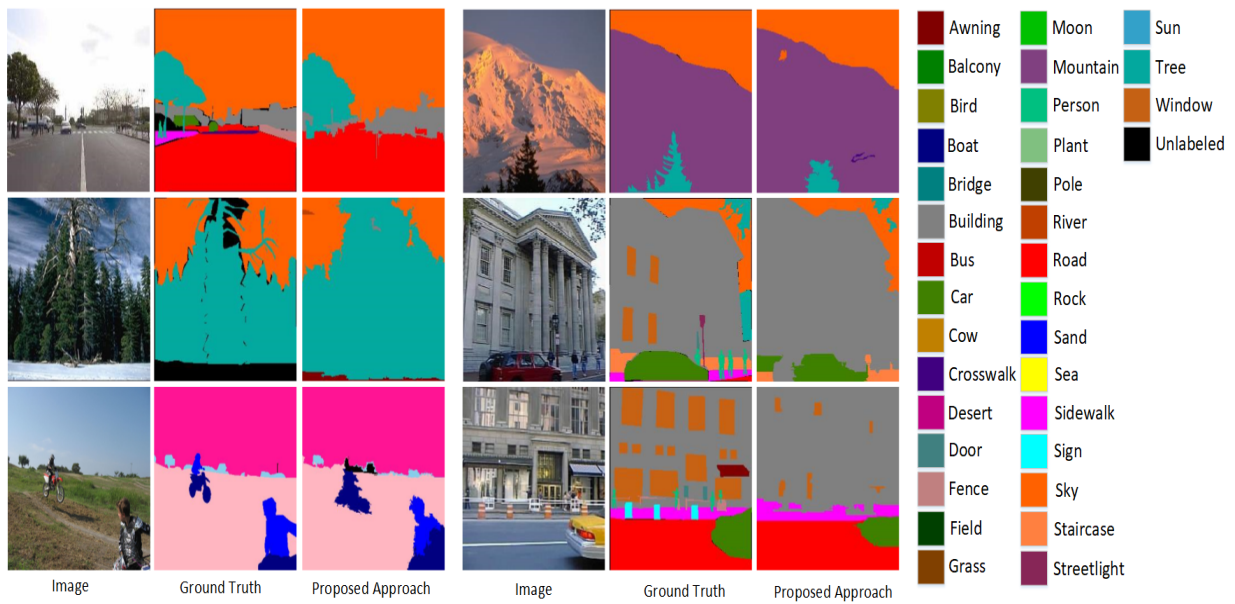

**Fig. 7.** Prediction results of sample images using the proposed approach on the SIFT Flow dataset (best viewed in colour).

2) *Performance comparisons on the PASCAL-Context dataset.* Table 4 compares the proposed approach with the best results reported by state-of-the-art approaches on the PASCAL-Context dataset. In addition to the global and class accuracies, we also report the PASCAL VOC Intersection-over-Union metric (IU) and Frequency Weighted IU (FW IU) which weights the IU of each class by its pixel frequency as defined in (Shelhamer et al., 2017). Observed from the table, our approach achieves a superior performance than state-of-the-art approaches, improving the recorded class accuracy from 64.7 to 74.1%, the recorded IU accuracy from 54.0 to 58.7% and the recorded FW IU accuracy from 53.0 to 61.3%. In addition, our approach also has one of the highest global accuracies. The results confirm state-of-the-art performance of the proposed non-parametric SCLP approach, which can generate more reliable test-specific contextual and visual features from big training data (note that the PASCAL-Context has more training images than the SIFT Flow dataset).

Fig. 8 visually displays the results of a set of image samples, confirming good overall performance obtained using the non-parametric SCLP approach. Similar to the results on the SIFT Flow dataset, there are some small misclassified image regions in some images. This indicates the necessity of further improving the proposed approach by incorporating 'noise' removal strategies such as enforcing a spatial constraint on the consistency of class labels in a larger neighbourhood of superpixels.

**Table 4**

Performance comparisons with state-of-the-art approaches on the PASCAL-Context dataset (59 classes).

| Approach | Global acc. | Class acc. | IU | FW IU |
|---|---|---|---|---|
| (Carreira et al., 2012) - O$_2$P | - | - | 18.1 | - |
| (Dai et al., 2015) - CFM | - | - | 34.4 | - |
| (Shelhamer et al., 2017) - FCN-32s | 65.5 | 49.1 | 36.7 | 50.9 |
| (Shelhamer et al., 2017) - FCN-16s | 66.9 | 51.3 | 38.4 | 52.3 |
| (Shelhamer et al., 2017) - FCN-8s | 67.5 | 52.3 | 39.1 | 53.0 |
| (Chen et al., 2018) - DeepLab | - | - | 45.7 | - |
| (Lin et al., 2019) - ZigZagNet | - | - | 52.1 | - |
| (Fu et al., 2019) - DANet | - | - | 52.6 | - |
| (Zhang et al., 2019) - CFNet | - | - | 54.0 | - |
| (Wu et al., 2019) - ResNet | 75.0 | 58.1 | 48.1 | - |
| (Ding et al., 2020) - CGBNet | 79.6 | 64.7 | 53.4 | - |
| Non-parametric SCLP | 75.4 | 74.1 | 58.7 | 61.3 |

Abbreviations: O$_2$P - Second Order Pooling, CFM - Convolutional Feature Masking, FCN - Fully Convolutional Network, DANet - Dual Attention Network, CFNet - Co-occurrent Feature Net, ResNet - Residual Network, CGBNet - Context-contrasted Gated Boundary Net.

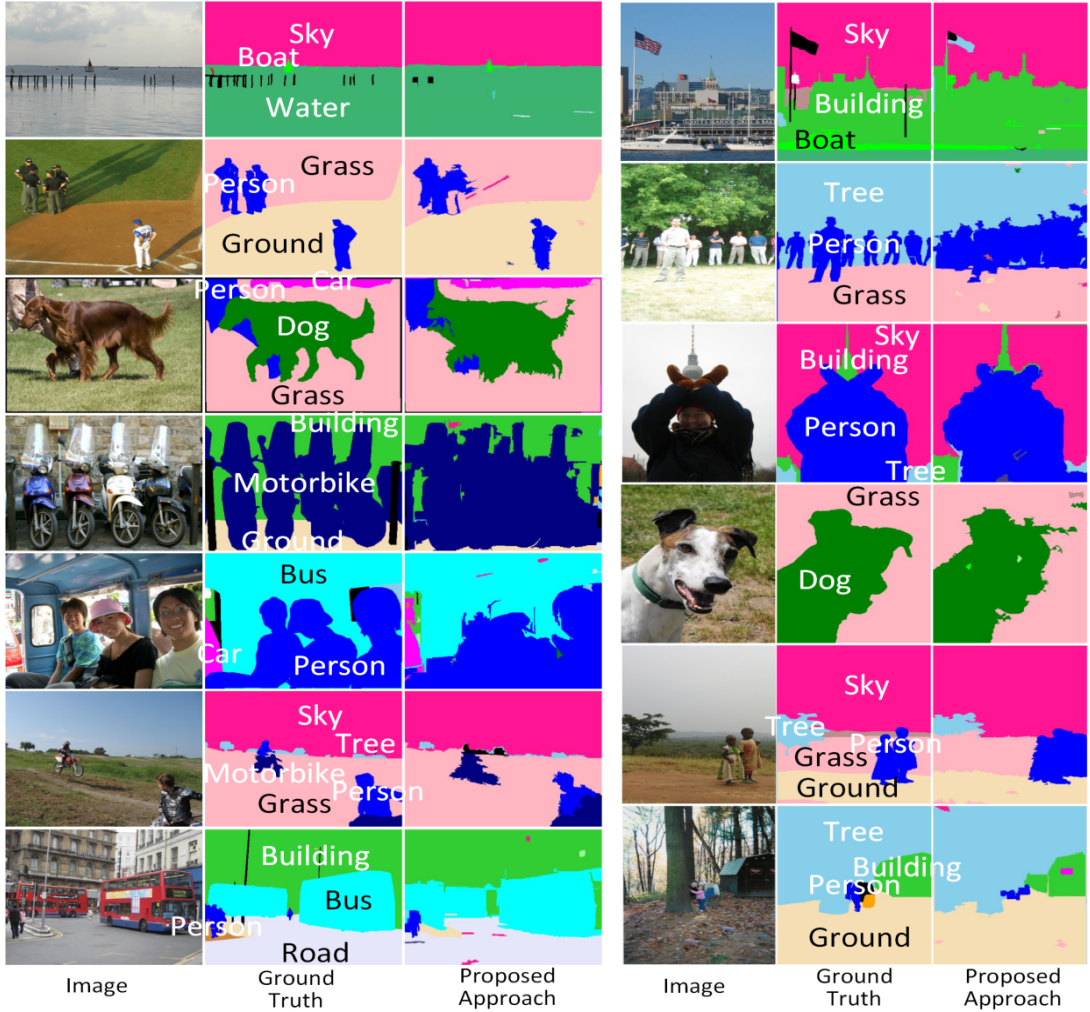

**Fig. 8.** Qualitative results on the PASCAL-Context dataset. There are some 'noise' in the classification results and they are due to the misclassification of small segmented superpixels (best viewed in colour).

## 5. Conclusions and Future Work

This paper presents a new non-parametric approach that utilizes SCLP contextual features for scene parsing on real-world data. A major novelty of the approach lies in the extraction of non-parametric SCLP contextual features for each query image from a small image subset retrieved from the training data. The image subset often shares similar scene content and spatial layouts to the query image. Therefore, the non-parametric SCLP is able to reflect more reliable and effective contextual information about inter-object correlations, specifically for every query image. Our experiments on the widely used SIFT Flow dataset show that the proposed approach achieves one of the top performance when compared with state-of-the-art approaches, and the integration of non-parametric SCLP with visual features leads to big increases on the accuracy than using visual features alone. The proposed approach also demonstrates state-of-the-art performance on a more recent and larger PASCAL-Context dataset. Our results also find that, in non-parametric approaches, retrieving a similar data subset for only rare classes can achieve higher global accuracy and require less computational time than retrieving a data subset for all classes, particularly for real-world datasets with heavily unequal pixel distributions between classes. One of our future directions is to investigate strategies to further refine the scene labelling results such as removing small misclassified regions by enforcing a spatial constraint on the consistency of class labels in a larger neighbourhood of superpixels, as well as incorporating multi-scale features.

The proposed approach has been successfully applied to the practice of automatic roadside fire risk prediction for the Queensland Department of Transport and Main Roads, Australia. It was used to segment roadside images into various object categories such as grass, tree, soil, road and building. From the segmented objects, further processing steps were performed to automatically predict the fire risk level, whereby actions can be taken to reduce the risk. The proposed approach can be potentially used to support many other practical applications such as autonomous driving, automatic hazardous object recognition, and image content retrieval.

**Acknowledgements**

We acknowledge the support from Australian Research Council and Queensland Department of Transport and Main Roads. This work was supported under the Australian Research Council's Linkage Projects funding scheme (project number LP140100939).